%% file: template-tex.tex
\documentclass{article}
\usepackage{spconf,amsmath,epsfig}
\usepackage{tikz}
\usepackage{multirow}
\usepackage{makecell}
\usepackage{pgfplots}
\pgfplotsset{compat=1.17}
\usetikzlibrary{shapes.geometric, arrows}

\usepackage[margin=1in]{geometry}

\tikzstyle{process} = [rectangle, rounded corners, minimum width=3cm, minimum height=1cm, text centered, draw=black, fill=blue!20]
\tikzstyle{arrow} = [thick,->,>=stealth]
\usepackage{booktabs}
\usepackage{subcaption} 
\usepackage{rotating}
\usepackage{tcolorbox}
\usepackage{float}
\tcbuselibrary{listingsutf8}
\providecommand{\doi}[1]{doi: {\footnotesize \href{http://dx.doi.org/#1}{\path{#1}}}}

\usepackage[pdftex=true,breaklinks=true,hidelinks=true,colorlinks=true,citecolor=blue]{hyperref}

\usepackage[square,numbers]{natbib}

\title{ASTREA: Introducing Agentic Intelligence for Orbital Thermal Autonomy}
\name{Alejandro D. Mousist}
\address{Thales Alenia Space, Tres Cantos, Spain}

\begin{document}

\ninept
\maketitle
%
\input{chapters/00_abstract}

%

\input{chapters/01_Introduction}

\input{chapters/02_LLMs}

\input{chapters/03_Experiment}

\input{chapters/04_Results}

\input{chapters/05_General_Discussion}

\input{chapters/06_Conclusions}

\section*{Acknowledgments}
I am grateful to my colleague Julian Cobos Aparicio for his careful review of the manuscript and for providing valuable editorial suggestions that enhanced its clarity. I also thank my friend and colleague David Alami for introducing me to the field of agentic AI, which inspired the conception of this paper.

\bibliographystyle{unsrt}
\bibliography{chapters/bibliography}

\small

\end{document}

%% file: chapters/00_Abstract.tex
\begin{abstract}
This paper presents ASTREA, the first agentic system executed on flight-heritage hardware (TRL 9) for autonomous spacecraft operations, with on-orbit operation aboard the International Space Station (ISS). Using thermal control as a representative use case, we integrate a resource-constrained Large Language Model (LLM) agent with a reinforcement learning controller in an asynchronous architecture tailored for space-qualified platforms.
Ground experiments show that LLM-guided supervision improves thermal stability and reduces violations, confirming the feasibility of combining semantic reasoning with adaptive control under hardware constraints. On-orbit validation aboard the ISS initially faced challenges due to inference latency misaligned with the rapid thermal cycles of Low Earth Orbit (LEO) satellites. Synchronization with the orbit length successfully surpassed the baseline with reduced violations, extended episode durations, and improved CPU utilization.
These findings demonstrate the potential for scalable agentic supervision architectures in future autonomous spacecraft.
\end{abstract}

%% file: chapters/01_Introduction.tex
\section{INTRODUCTION}
\label{sec:intro}


\begingroup
\setlength{\parskip}{0.5em}   
\setlength{\parindent}{1em} 

Recent advances in robotics and artificial intelligence have accelerated the possibilities for autonomy, particularly in space exploration missions. However, AI could potentially transform the way lunar missions and Earth-orbiting missions are operated and monitored, especially in scenarios requiring real-time decision-making as an integral part of mission operations. This technological evolution represents a paradigmatic shift toward space systems capable of operating with minimal human intervention, particularly critical in environments where communication delays preclude direct ground supervision.

Numerous autonomous control initiatives have been documented, ranging from assistance in traditional subsystems to autonomous capture and landing mission planning. These developments have demonstrated the technical feasibility of delegating critical decisions to automated systems, establishing the foundation for the next generation of fully autonomous space missions.

Recently, the emergence of Large Language Models (LLMs) and their integration into the new agentic paradigm have added a new dimension to autonomous decision-making capabilities, enabling the interpretation of nuances, ambiguous contexts, and non-deterministic decision-making based on objectives or historical data, approaching human-like behavior in decision processes. This semantic reasoning capability opens previously unimaginable possibilities for space autonomy, where contextual interpretation and adaptability are fundamental to mission success.

However, these large models also require substantial hardware resources in their most powerful versions, which conflicts with embedded environments that must maintain radiation tolerance and operate under severe constraints regarding power consumption, size, and temperature. This dichotomy between desired capabilities and the physical limitations of space hardware represents one of the primary challenges for the practical implementation of agentic systems in flight applications.

Thermal control constitutes a critical subsystem that perfectly exemplifies these challenges, as it must maintain the operational integrity of all electronic components while managing limited computational resources in real-time. The inherent complexity of balancing variable thermal loads with dynamic processing demands requires a level of decision-making sophistication that has traditionally relied on pre-programmed rules and ground-based supervision.
This work addresses this gap by implementing an agentic system on flight-qualified hardware (TRL 9) currently operating in the IMAGIN-e\cite{callejo2023imagin} mission, where an LLM agent provides contextual recommendations for thermal control parameters while maintaining operational independence for safety-critical applications. Beyond ground validation, the system was also deployed in on-orbit experiments aboard the International Space Station (ISS), representing, to the best of our knowledge, the first demonstration of an agentic LLM-based supervisory system executed in a real flight environment.

This system, named \textbf{ASTREA} (\textbf{A}gentic \textbf{S}ystem for \textbf{T}hermal \textbf{R}egulation and \textbf{E}mbedded \textbf{A}daptation), demonstrates how advanced reasoning capabilities can be integrated into real space systems, overcoming hardware limitations through a hybrid design that combines the efficiency of reinforcement learning with the interpretability of linguistic models.
The experimental validation of this approach not only confirms the technical feasibility of agentic systems in real space environments but also establishes a precedent for future implementations where intelligent autonomy is fundamental to mission success. The obtained results suggest that careful integration of advanced AI technologies into flight hardware can provide sophisticated decision-making capabilities without compromising the reliability required in critical space applications.

The paper proceeds as follows: Section 2 reviews the state of the art in LLM-based intelligence autonomy for space exploration. Section 3 details the experimental design and implementation of ASTREA, including the LLM and RL agents, hardware platform, and evaluation environments. Section 4 presents the results from ground and on-orbit experiments, along with a comparative analysis against the baseline. Section 5 discusses the implications for future space autonomy, and Section 6 concludes with key findings and directions for future work.

\endgroup

%% file: chapters/02_LLMs.tex
\section{State of the art in Intelligence Autonomy based on LLMs}
\subsection{LLMs in Space}
Space Llama \cite{boozallen2025space} marks a significant milestone as the first open-source LLM (Llama 3.2) deployed and operated aboard the International Space Station (ISS). 
Developed by Booz Allen and Meta, Space Llama is engineered to assist astronauts in scientific and technical tasks, including predictive maintenance, autonomous access to documentation, and the replacement of physical manuals. 

However, its design lacks agentic behavior, as it was not intended for autonomous mission control but rather for onboard human use.

\subsection{Agentic LLM systems for Space Exploration}
Recent efforts to integrate Large Language Models (LLMs) into spacecraft autonomy have primarily focused on ground-based experimentation with agentic architectures. In LLMSat \cite{maranto2024llmsatlargelanguagemodelbased}, an LLM is proposed as a high-level control system for spacecraft, functioning as a goal-driven autonomous agent overseeing general mission operations. The study explores the LLM’s role as “Pilot,” responsible for strategic decision-making and direct vehicle control. Evaluated in a simulated software environment, the system exhibited mixed cognitive performance, with a marked decline as mission complexity increased.

Similarly, AI Space Cortex \cite{touma2025ai} introduces a centralized and explainable autonomous framework designed for future planetary missions in extreme environments, such as icy moons. This system leverages LLMs for semantic reasoning, environmental perception, autonomous site selection, and dynamic strategy adaptation—operating under distinct exploration modes ("Conservative," "Scientific Curiosity," and "Adventurous"). Validated on terrestrial testbeds, AI Space Cortex highlights the relevance of resilient and interpretable autonomy in domains with severe latency and operational uncertainty.

\subsection{Combined LLM and RL Agents in Hybrid Systems}
Recent advances in autonomous systems have explored the integration of heterogeneous agents that combine the semantic reasoning capabilities of LLM-based agents with the adaptive control mechanisms offered by Reinforcement Learning (RL)-based agents. These hybrid architectures seek to leverage the complementary strengths of each approach: LLM-agents excel in high-level deliberation, natural language interaction, and context-aware decision support, while RL agents operate effectively in dynamic environments through policy optimization and reward-based learning. In space applications, such architectures offer promising pathways for constructing resilient and interpretable autonomous systems capable of responding to complex operational scenarios, particularly where communication delays and environmental uncertainty impose constraints on centralized supervision.

Schoepp S. et al.\cite{schoepp2025evolving} identify three ways of integrating LLMs in RL agents, distinguishing three primary roles: agent, where the LLM directly acts as the policy making decisions; planner, where the LLM decomposes complex tasks into subtasks or intermediate goals; and reward model, where the LLM helps generate or evaluate reward signals. Following this taxonomy, Navarro T. et al. \cite{Navarro2024-tl} propose the use of an LLM as a reward generator within a reinforcement learning framework designed to assist in planetary landing tasks, demonstrating promising results in the practical application of LLMs to complex, high-stakes decision-making environments.

However, Chen Z. et al.\cite{chen2025hcrmpllmhintedcontextualreinforcement} explicitly address the problem of hallucinations generated by LLMs. They note that LLM-dominated frameworks—where the language model plays a central role in decision-making—are vulnerable to semantic errors that may compromise the safety of RL agents, even when using state-of-the-art LLMs. To mitigate this risk, they advocate for maintaining a relative independence between the LLM and the RL agent.

Building on this last proposal, the present study examines the integration of a reinforcement learning (RL) agent for onboard satellite thermal control with a language model (LLM) agent. In this system, the LLM provides suggestions for adjusting the $\alpha$ parameter of the RL model (Soft Actor-Critic) which governs its exploratory behavior. The RL agent incorporates these recommendations while maintaining operational independence from the LLM.

The constrained hardware environment of space applications necessitates the use of smaller LLM models, usually conceived for edge computing, and also introduces significant inference latency, providing additional justification for excluding the LLM from the real-time control loop. Instead, the LLM operates as a higher-level supervisory component that analyzes extended time windows and provides medium-term strategic recommendations, while the RL agent handles immediate thermal control decisions.
This hybrid architecture forms the foundation of ASTREA, a system designed to combine the semantic reasoning of LLMs with the adaptive control of RL agents in constrained onboard environments. ASTREA operates asynchronously, enabling strategic supervision without compromising real-time responsiveness.



%% file: chapters/03_Experiment.tex
\section{Experimental Design and Implementation}
\label{sec:methodology}

\subsection{General design of the experiment}
The experimental setup described in this study corresponds to ASTREA, a hybrid agentic system composed of two core modules: a reinforcement learning agent responsible for thermal regulation, and a language model agent that provides semantic recommendations.

\begin{figure}[ht]
  \centering
  \includegraphics[width=\columnwidth]{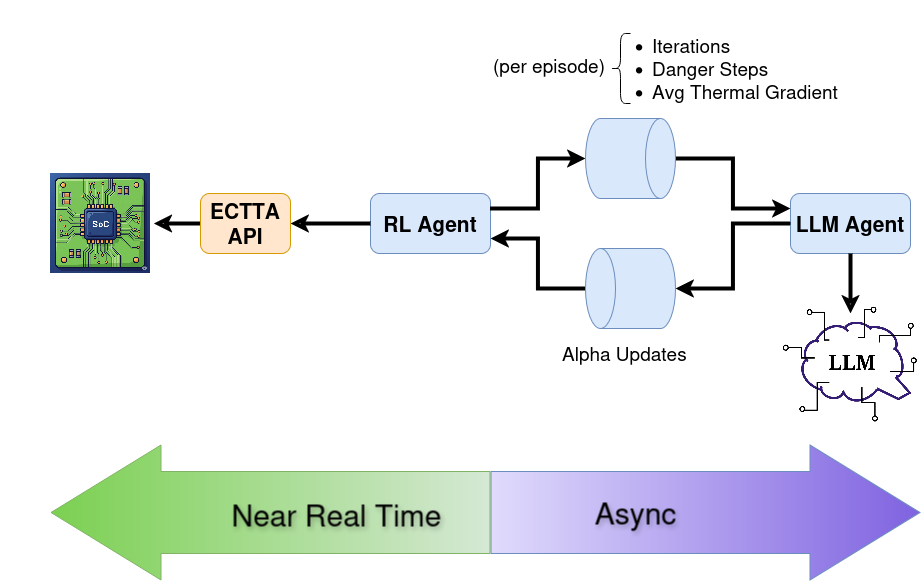}
  \caption{Schematic diagram of the experiment. The RL Agent send the iteration count, how many of that iterations have been near the threshold ($<1~^\circ\mathrm{C}$) and the average thermal gradient of the last episode. The LLM Agent answers asynchronously the new alpha suggestion}
  \label{fig:schema}
\end{figure}

Figure \ref{fig:schema} illustrates the overall experimental design. The RL-Agent continuously monitors the board temperature and promptly adjusts resource availability on the edge node. At the end of each episode, it transmits a summary to the LLM-Agent via an asynchronous queue. This summary includes the total number of iterations, the count of iterations where the peak temperature recorded by any sensor is within one degree of the critical threshold, and the average thermal gradient observed throughout the episode.

Meanwhile, the LLM-Agent periodically checks the episode summary queue and stores the data locally. Once a predefined time window elapses, all stored summaries are processed and fed into an LLM to generate entropy coefficient ($\alpha$) value recommendations for the SAC model used by the RL-Agent.

These recommendations are transmitted back to the RL-Agent through a separate queue. During each iteration, the RL-Agent checks for updated $\alpha$ values, and if a new recommendation is detected, it is immediately applied to the model.

This asynchronous communication architecture enables the coexistence of two complementary agentic modalities: a real-time RL-Agent focused on immediate thermal regulation, and a higher-latency LLM-Agent capable of deeper contextual analysis. By decoupling their operational timelines, the system leverages rapid responsiveness alongside strategic modulation, enhancing both short-term control and long-term adaptability.


\subsection{LLM Agent}
There is only one LLM-Agent present in the system and its only task is to decide which is the best entropy coefficient ($\alpha$) of the RL-Agent, which performs the thermal control of the system. 

The LLM-Agent leverages an unrefined general-purpose LLM to address the target problem. The agent's design capitalizes on the LLM's broad knowledge base, guiding its reasoning through carefully crafted prompt engineering. While this approach facilitates rapid implementation, it requires the agent to abstract the problem appropriately, pre-compute values that provide direct and relevant information for decision-making, and constrain the model's scope and operational domain to ensure the problem remains manageable.

\begin{figure}[t]
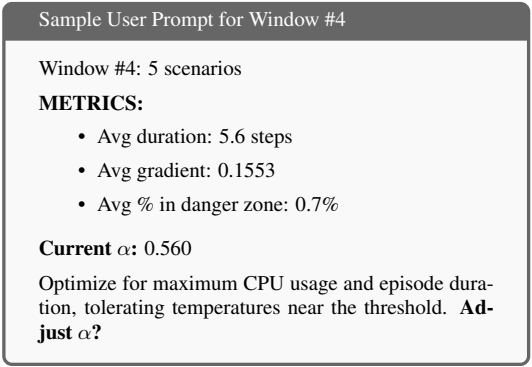

\centering
\resizebox{0.9\columnwidth}{!}{
    \begin{tcolorbox}[title=Sample User Prompt for Window \#4, colback=gray!5!white, colframe=gray!80!black]
        Window \#4: 5 scenarios
        
        \vspace{0.5em}
        
        \textbf{METRICS:}
        \begin{itemize}
            \item Avg duration: 5.6 steps
            \item Avg gradient: 0.1553
            \item Avg \% in danger zone: 0.7\%
        \end{itemize}
        \vspace{0.5em}
        \textbf{Current $\alpha$:} 0.560
        \vspace{0.5em}
        
        Optimize for maximum CPU usage and episode duration, tolerating temperatures near the threshold. \textbf{Adjust $\alpha$?}
    \end{tcolorbox}
    }
\caption{User prompt used by the LLM-Agent to ask the language model for $\alpha$ updates}
\label{fig:user-prompt}
\end{figure}

Given the above, the LLM-Agent employs a static system prompt (see Fig~\ref{fig:system-prompt}) to configure the behavior of the language model, and a dynamic user prompt (see Fig~\ref{fig:user-prompt}) to request recommendations for updating the $\alpha$ coefficient. The generation of recommendations relies on the most recent time window of episode summaries, that includes average duration, average gradient and average of percentage steps in danger zone, along with the current $\alpha$ value.

The reason for sending this information to the LLM is to provide information on the performance of the RL-agent over the time window so that the LLM-agent can apply the rules (defined in the system prompt) and, using recommended tools, determine  an $\alpha$ recommendation.

Initial preliminary experiments revealed LLM response times of approximately 10 minutes in certain instances. Considering these durations, alongside the time required for the RL model to adapt to new alpha values and cool-down periods between episodes, a time window was established for collecting episode summaries before triggering alpha value recommendations. 

The specific window size was selected based on the operational environment to ensure sufficient data collection for meaningful LLM recommendations while maintaining reasonable system responsiveness for thermal control applications.

\begin{figure}[t]
\centering
\resizebox{0.9\columnwidth}{!}{
    \begin{tcolorbox}[title=Sample System Prompt for a Threshold of 60$^\circ$C, colback=gray!5!white, colframe=gray!80!black]
        \small
        You optimize SAC entropy coefficient ($\alpha$) to maximize CPU usage and episode duration for thermal control. It's optimal to keep temperature just below the threshold.\\
        \textbf{THERMAL LIMIT:} 60$^\circ$C (critical threshold)

        \vspace{0.5em}
        \textbf{UTILIZATION PATTERNS:}
        
        \begin{itemize}
            \item Low utilization (under-utilization): Short scenarios ($<$10 steps) or low \% in danger zone ($<$20\%) – indicates low CPU push or violations too quick.
            \item Medium utilization: Medium scenarios (10--60 steps) or medium \% in danger zone (20--60\%)
            \item High utilization (optimal utilization): Long scenarios ($>$60 steps) or high \% in danger zone ($>$60\%) – good hugging near limit without violations.
        \end{itemize}
        Additional context: Avg thermal gradient informs heating rate (high $>0.1$ = fast change, use for decision nuance but not primary classification).

        \vspace{0.5em}
        \textbf{$\alpha$ STRATEGY:}
        \begin{itemize}
            \item Low utilization $\rightarrow$ INCREASE $\alpha$ (0.4--0.8): More exploration to find higher CPU strategies without violations.
            \item Medium utilization $\rightarrow$ MODERATE $\alpha$ (0.2--0.4): Balanced approach.
            \item High utilization $\rightarrow$ DECREASE $\alpha$ (0.05--0.2): Exploit proven high-utilization policies.
            \item Mixed/uncertain $\rightarrow$ Keep current or slightly adjust ($\pm$0.05).
        \end{itemize}

        \vspace{0.5em}
        Consider the average percentage of steps in the danger zone ($\geq$59$^\circ$C) when assessing risk. A high percentage (without violations) indicates optimal max CPU near the limit; treat as high utilization if episodes long. Low percentage suggests under-utilization.

        \vspace{0.5em}
        \textbf{Briefly justify $\alpha$ choice in 1 line, then call a tool. Tool use is mandatory.}

    \end{tcolorbox}
    }
\caption{System prompt used by the LLM-Agent to configure the behavior of the language model}
\label{fig:system-prompt}
\end{figure}

The LLM-agent uses four tools for making the recommendations, which are:
\begin{itemize}
    \item \emph{Increase exploration}: Increase exploration with alphas in the upper range ($\alpha \in [0.4, 0.8]$)
    \item \emph{Moderate exploration}: Set exploration in a more moderate range ($\alpha \in [0.2, 0.4]$)
    \item \emph{Decrease exploration}: Decrease exploration and exploit proven efficient policies with alphas in a lower range ($\alpha \in [0.05, 0.2]$)
    \item \emph{Keep alpha}: Keep current value of alpha.
    \item \emph{Reset alpha}: Sets alpha to a default value ($\alpha=0.2$).
\end{itemize}

For the agent implementation, the provider-agnostic framework developed by OpenAI, known as the \emph{OpenAI Agents SDK} \cite{openai_agents_2024}, was chosen due to its simplicity, lightweight design, and suitability for the intended use case.

\subsubsection{Large Language Model}
The LLM used by the LLM-agent is a specific variant of Qwen2.5\cite{qwen2, qwen2.5}, featuring 1.54 billion parameters and quantized to 4 bits. While Qwen2.5 includes multiple model sizes and quantization levels, this configuration was selected to balance computational efficiency and reasoning performance. 

Although detailed benchmarks for 4-bit quantized versions of Qwen2.5 are limited, experimental results from the Qwen3 family suggest that models maintain competitive performance at 4-bit precision, with more pronounced degradation only observed at 3-bit quantization or lower \cite{zheng2025empiricalstudyqwen3quantization}. In the context of the present use case, the selected Qwen2.5 variant has demonstrated reasoning capabilities comparable to higher-precision alternatives.

On-device inference is carried out using Llama.cpp \cite{llama_cpp}, a lightweight and hardware-efficient inference framework that delivers state-of-the-art performance with minimal setup. The server implementation is compatible with the OpenAI API, enabling seamless integration with the agentic framework.

\subsection{RL Agent}
\label{sec:rl}

The RL-agent used in this work was originally presented in \cite{mousist2024autonomous}, where its design, training, and validation are detailed in the context of autonomous thermal control for small satellites. The agent is based on the Soft Actor-Critic (SAC) algorithm and was trained to regulate the satellite thermal system by managing processing resources while the cores are subjected to continuous computational load generated via the \emph{stress-ng} utility.

In this work, the reward function has been revised to improve convergence and better balance performance with thermal safety. Specifically, a base survival reward and a thermal safety bonus were introduced, encouraging the agent to maintain safe operating margins while optimizing CPU usage. 

In addition, a new observation feature called \emph{danger ratio} has been added, which quantifies the proportion of thermal sensors operating within 10\% of their defined threshold. This provides the agent with a real-time measure of proximity to thermal limits, enhancing its ability to anticipate and mitigate thermal risks.

\begin{figure}[H]
  \centering
  \includegraphics[width=\columnwidth]{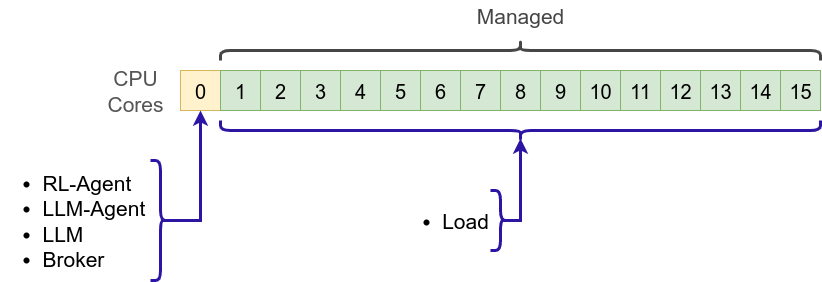}
  \caption{Core-0 is reserved for the agentic thermal control system and the other 15 cores are under heavy load and managed by the agents.}
  \label{fig:cores_agentic}
\end{figure}

In contrast to the configuration described in the previous study \cite{mousist2024autonomous}, where the agent controlled all cores, the current setup reserves core 0 for system management tasks, operating at maximum frequency. The RL-agent now manages the remaining 15 cores, dynamically adjusting both their frequency and power state. Figure \ref{fig:cores_agentic} illustrates this distribution of the CPU cores in the experiment.

\subsection{Hardware Platform and Computational Constraints}
The experimental setup was deployed on a flight-qualified System-on-Chip (SoC) featuring a 64-bit ARM architecture with 16 Cortex-A72 cores. The platform includes 16 GB of LPDDR4 memory and operates within a dynamic frequency range between 1.0 GHz and 2.0 GHz. No dedicated hardware accelerators (e.g., GPU, NPU) are available, which imposes strict limitations on real-time inference and parallel processing.

To ensure deterministic behavior and isolate agentic control from concurrent tasks, core 0 was reserved exclusively for ASTREA’s orchestration layer. The remaining 15 cores were subjected to continuous computational stress, simulating realistic onboard conditions. This architectural constraint directly influenced the design choices described in subsequent sections, including the adoption of a quantized LLM agent and the asynchronous execution model.

\subsection{Experimental Environments: Ground Lab and ISS Deployment}
To validate ASTREA under realistic operational conditions, the system was evaluated in two distinct environments: a semi-controlled ground laboratory and an external payload platform aboard the International Space Station (ISS). Each environment imposed unique constraints that shaped both the system’s behavior and its architectural design.

\subsubsection{Ground Laboratory Setup}
\label{sec:ground}
The ground experiments were conducted in a mini-rack configuration equipped with active cooling via fans. Ambient temperature was regulated for 10 hours during the day but remained unconditioned overnight, introducing natural thermal fluctuations. The SoC operated under continuous computational stress, with 15 cores fully loaded and core 0 reserved for agentic orchestration. For this environment, a 60-minute time window was established for collecting episode summaries before triggering alpha value recommendations from the LLM Agent.

This setup provided a stable yet dynamic environment for evaluating ASTREA’s semantic supervision capabilities under constrained but predictable conditions. The laboratory configuration allowed for fine-grained control over thermal and computational variables, facilitating reproducible experimentation.

\subsubsection{On-Orbit Deployment (ISS)}
\label{sec:iss}
The second evaluation environment was an external payload mounted on the Columbus module of the ISS. This orbital configuration exposed the system to cyclical thermal transitions driven by the station’s 90-minute orbit—approximately 45 minutes of direct solar exposure followed by 45 minutes of eclipse.

The payload included a passive heat pipe for thermal mitigation, but no active cooling mechanisms. These constraints introduced challenges for real-time inference and agentic responsiveness, particularly in relation to the rapid environmental shifts inherent to low Earth orbit.

\subsection{Baseline and Evaluation Metrics}
\subsubsection{Baseline: SAC with adaptive $\alpha$}

\begin{figure}[H]
  \centering
  \includegraphics[width=\columnwidth]{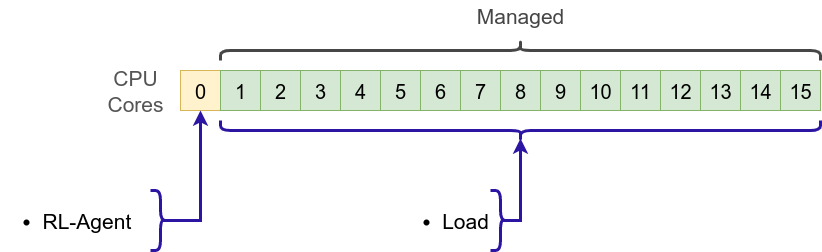}
  \caption{Only the RL-agent operates in the core-0 for performing thermal control. As in the agentic version, the load is applied on the other cores.}
  \label{fig:cores_rl}
\end{figure}

The baseline system employs the same RL-agent architecture described in Section \ref{sec:rl}, but uses the default adaptive $\alpha$ scheduling mechanism from Stable Baselines3 \cite{stable-baselines3}. This implementation automatically adjusts the entropy coefficient during training according to the internal heuristics of the algorithm, without external intervention or episode-level performance feedback (Represented in Fig. \ref{fig:cores_rl}).

Unlike the proposed LLM-guided approach, the baseline system does not collect thermal episode metrics, generate performance summaries, or receive contextual parameter recommendations based on thermal safety patterns.

\subsubsection{Evaluation Metrics}
The following metrics are used to compare the baseline and proposed systems over a 24 hours evaluation period:
\begin{itemize}
    \item \emph{Number of thermal violations}: Count of instances where any thermal sensor exceeds the temperature threshold during the evaluation period.  
    \item \emph{Average episode duration}: Mean duration (in time steps) of the thermal control episodes, measured over the evaluation period.
    \item \emph{CPU utilization efficiency}: Inverse of remaining computational capacity, accounting for frequency scaling where high-frequency cores contribute fully to utilization, low-frequency cores contribute partially, and inactive cores contribute zero. 
\end{itemize}

%% file: chapters/04_Results.tex
\section{Results and Comparative Analysis}
\label{sec:results}
\subsection{Ground Experiments}
The performance of ASTREA was first evaluated against a baseline RL-only controller, under semi-controlled laboratory conditions (See \ref{sec:ground}). Within this environment, ASTREA achieved longer episode durations and significantly fewer thermal violations, particularly during the initial operational phase. These improvements underscore the value of semantic modulation in enhancing RL agent behavior under constrained and partially variable conditions.

\begin{figure}[H]
    \begin{center}
    \begin{tikzpicture}
        \begin{axis}[
            width=\columnwidth,
            height=6cm,
            xmin=0,
            ymin=0,
            axis lines=left,
            enlargelimits=false,
            xlabel={Iteration},
            ylabel={Episode Length Mean},
            legend style={at={(0.05,0.95)}, anchor=north west, font=\scriptsize}
        ]
        \addplot[
            very thick,            
            green!70!black,                  
            mark=none              
        ] coordinates {
            (17,17) (26,9) (39,13) (42,10.5)
            (58,11.6) (102,17) (134,19.1)
            (160,20) (195,21.7) (334,33.4)
            (381,34.6) (404,33.7) (531,40.8)
            (1642,117) (1671,111) (1813,113)
            (1947,115) (1968,109) (2700,142)
        };
        \addlegendentry{Agentic}

        \addplot[
            very thick,            
            blue,                  
            mark=none              
        ] coordinates {
            (32,32) (41,13.3) (60,20) (76,19)
            (106,21.2) (124,20.7) (160,22.9) (174,21.8)
            (186,20.7) (207,20.7) (259,23.5) (334,27.8)
            (429,33) (468,33.4) (527,35.1) (586,36.6)
            (607,35.7) (629,34.9) (691,36.4) (741,37)
            (750,35.7) (778,35.4) (782,34) (796,33.2)
            (828,33.1) (842,32.4) (910,33.7) (917,32.8)
            (926,31.9) (962,32.1) (978,31.5) (998,31.2)
            (1062,32.2) (1076,31.6) (1115,31.9) (1124,31.2)
            (1259,34) (1283,33.8) (1412,36.2) (1440,36)
            (1813,44.2) (1947,46.4) (1968,45.8) (2015,45.8)
            (2105,46.8) (2233,48.5) (2360,50.2) (2553,53.2)
            (2571,52.5) (2700,54) (2726,43.2)
        };
        \addlegendentry{Baseline}
        
        \addplot[
            red, dashed, very thick
        ] coordinates {(404,0) (404,150)};

        \addplot[
            red, dashed, very thick
        ] coordinates {(1671,0) (1671,150)};

        \addplot[
            red, dashed, very thick
        ] coordinates {(1968,0) (1968,150)};
        
        \addplot[
            red, dashed, very thick
        ] coordinates {(0,0) (1,0)};
        \addlegendentry{Alpha updates}

        \end{axis}
    \end{tikzpicture}
    \end{center}
    \caption{Episode length mean for baseline and agentic executions. The dashed red lines show the iterations in which the LLM-agent gave recommendations for $\alpha$ parameter in the agentic implementation.}
    \label{fig:alpha_update}
\end{figure}
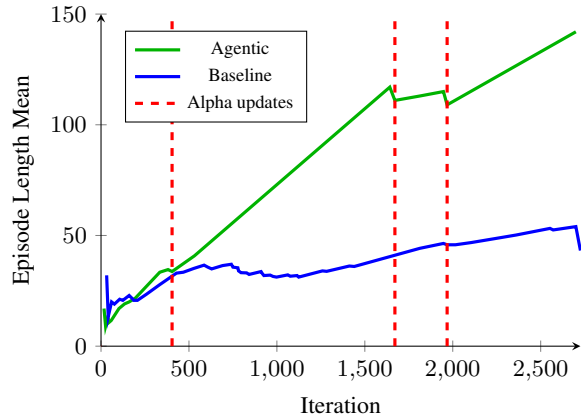

Initially, both the agentic and baseline systems exhibited comparable behavior. However, following the first entropy coefficient ($\alpha$) adjustment recommended by the LLM-based agent, the agentic implementation transitioned into markedly longer episodes of activity—a 67.2\% increase in average duration during the first 4 hours compared to the baseline. In contrast, the baseline system maintained a trajectory of gradual, incremental improvement. This divergence in behavior is illustrated in Figure~\ref{fig:alpha_update}, which highlights the influence of LLM-guided parameter modulation on the reinforcement learning (RL) agent’s operational dynamics.

Table~\ref{table:agentic_baseline} presents a comparative analysis of both systems across three key metrics: average episode duration, number of thermal violations, and average CPU utilization. The agentic system not only sustained longer episodes but also achieved a 58.5\% reduction in thermal violations during the early phase, and a 42.1\% reduction over 24 hours, indicating significantly improved thermal compliance. These gains were achieved with minimal impact on CPU usage, which remained nearly constant—only a 1.9\% increase during the first 4 hours and a 1.5\% decrease over the full 24-hour period.

Longer episodes are indicative of improved thermal stability, as the agent successfully maintains system temperature below the critical threshold for extended periods. A reduced number of thermal violations further reflects enhanced adherence to thermal constraints. These metrics are inherently interdependent, given that episodes terminate upon exceeding the thermal limit—thus, longer durations typically correlate with fewer violations.

It is important to note that the RL agent is also tasked with maximizing CPU usage without triggering thermal violations. This dual objective prevents the agent from simply throttling performance to extend episode duration, ensuring a balanced optimization strategy.

During the initial 4-hour period, the agentic system demonstrated substantial improvements in both episode duration and thermal compliance, likely due to the exploratory strategies introduced by the LLM-agent. Over the full 24-hour period, while the difference in episode duration narrowed to just 5.2\%, the agentic system maintained superior thermal performance, suggesting that LLM-guided modulation yields early advantages that persist in long-term stability.
This early advantage may also contribute to the system’s ability to dynamically adapt to shifting operational conditions, further reinforcing the role of the LLM-agent as a flexible and context-aware modulator in unpredictable space environments.

\begin{table}[h]
\caption{Experiment results on ground in early stages and after 1 day of execution.}
\label{table:agentic_baseline}
\centering
\small
\resizebox{\columnwidth}{!}{
\begin{tabular}{|l|l|r|r|}
    \hline \hline
    \multicolumn{1}{|c|}{\textbf{Metric}} & 
    \multicolumn{1}{c|}{\textbf{Duration}} & 
    \multicolumn{1}{c|}{\makecell{\textbf{Baseline}\\ (Mean $\pm$ Std. Dev)}} & 
    \multicolumn{1}{c|}{\makecell{\textbf{Agentic}\\ (Mean $\pm$ Std. Dev)}} \\
    \hline
    \multirow{2}{*}{Avg. Episode Duration} & First 4h & $47.17 \pm 18.15$ & $\mathbf{78.83 \pm 11.33}$ \\
                                              & 24h      & $135.24 \pm 32.94$ & $\mathbf{142.29 \pm 8.06}$ \\
    \hline
    \multirow{2}{*}{Thermal Violations}       & First 4h & $39.33 \pm 9.29$ & $\mathbf{16.33 \pm 2.08}$ \\
                                              & 24h      & $88.67 \pm 20.50$ & $\mathbf{51.33 \pm 4.04}$ \\
    \hline
    \multirow{2}{*}{Avg. CPU Usage (\%)}           & First 4h & $25.81 \pm 5.00$ & $\mathbf{26.30 \pm 2.56}$ \\
                                              & 24h      & $\mathbf{16.49 \pm 5.57}$ & $16.24 \pm 4.46$ \\
    \hline \hline
\end{tabular}
}
\end{table}

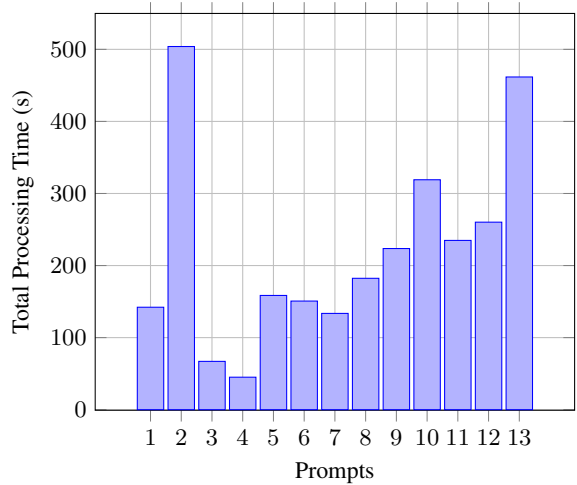
\begin{figure}[H]
    \begin{center}
        \begin{tikzpicture}
            \begin{axis}[
                ybar,
                width=\columnwidth,,
                ylabel={Total Processing Time (s)},
                xlabel={Prompts},
                xtick=data,
                enlarge x limits=0.15,
                grid=major,
            ]
            \addplot coordinates {
                (1,142.16652)
                (2,503.93877)
                (3,67.04244)
                (4,45.08423)
                (5,158.52138)
                (6,150.72553)
                (7,133.59947)
                (8,182.3077)
                (9,223.62077)
                (10,319.04848)
                (11,234.8831)
                (12,260.19935)
                (13,461.65709)
            }; 
            \end{axis}

        \end{tikzpicture}
    \end{center}
    \caption{Total processing time (latency) reported by the LLM for twelve individual prompt executions.}
    \label{fig:latency}
\end{figure}

Finally, as shown in Figure~\ref{fig:latency}, LLM-agent response times ranged from approximately 40 seconds to over 8 minutes, depending on prompt size and the operational status of core 0. These latencies clearly indicate that, within the present configuration, incorporating the LLM-agent into the real-time control loop of the RL-agent is not feasible. Such delays in decision-making could pose a thermal risk to the systems under supervision, particularly during high-stress operational scenarios.

It is important to note, however, that in this experiment, all LLM processing was confined to core 0, thereby precluding any parallelization strategies typically available when leveraging multiple cores. This architectural constraint significantly contributes to the observed latency. 

\subsection{On-Orbit Experiments (ISS)}
ASTREA was subsequently deployed aboard the International Space Station (ISS) to validate its feasibility in a real operational environment (see Section~\ref{sec:iss}). In the flight experiments, the maximum temperature was established at 57°C, a safe operational limit for the onboard equipment, with all experiments conducted during September 2025.

Two on-orbit experiments were conducted on this platform: a short-cycle configuration, and a longer one aligned with the orbital period.

\subsubsection{Short-cycle experiment}
For the first on-orbit configuration, the analysis window of the LLM agent was set to a conservative 15 minutes. This duration was selected as a prudential minimum considering the system’s measured inference latency: some queries required up to 8 minutes for completion (see Figure~\ref{fig:latency}). Although a 10-minute window could have more closely matched the shortest stable orbital thermal cycles, the additional margin was intentionally introduced to ensure timing robustness and operational safety. The goal was to shorten the reasoning cycle as much as safely possible, considering the LLM’s inference latency and the fast thermal variations experienced in orbit.

In this configuration, the agentic system exhibited an increase in the number of thermal violations and also produced shorter episode durations than the RL-only baseline. CPU utilization decreased, reflecting inefficient resource allocation under these dynamic conditions (Table~\ref{table:iss_results}).

\begin{table}[h]
\caption{Representative results from on-orbit short-cycle experiments aboard the ISS.}
\label{table:iss_results}
\centering
\small
\resizebox{\columnwidth}{!}{
\begin{tabular}{|l|r|r|}
    \hline \hline
    \multicolumn{1}{|c|}{\textbf{Metric}} & 
    \multicolumn{1}{c|}{\makecell{\textbf{Baseline}\\ (Mean $\pm$ Std. Dev)}} & 
    \multicolumn{1}{c|}{\makecell{\textbf{Agentic}\\ (Mean $\pm$ Std. Dev)}} \\
    \hline
    Avg. Episode Duration & $\mathbf{449.77 \pm 295.69}$ & $363.76 \pm 190.23$ \\
    \hline
    Thermal Violations    & $\mathbf{66 \pm 24.25}$ & $82 \pm 32.74$ \\
    \hline
    Avg. CPU Usage (\%)   & $\mathbf{40.17 \pm 17.70}$ & $22.86 \pm 03.05$ \\
    \hline \hline
\end{tabular}
}
\end{table}

The degraded performance profile can be partially explained by a temporal mismatch between the LLM decision cycle and the orbital thermal dynamics, combined with the stabilizing effect of the agentic policy on thermal control. Unlike terrestrial experiments, where ambient temperature evolves gradually and predictably, the ISS undergoes rapid and extreme thermal transitions during its 90-minute orbit—approximately 45 minutes of direct solar exposure followed by 45 minutes of eclipse. These sharp transitions often occurred while the LLM was still processing its previous cycle, rendering some $\alpha$ recommendations outdated at the moment of application.

As a result, certain updates introduced exploratory behavior during phases that required more conservative exploitation, while in other cases the agentic modulation successfully reduced the magnitude and frequency of thermal violations.

These observations motivated a second on-orbit configuration designed to align the reasoning cycle with the orbital period, aiming to test whether temporal synchronization could overcome the limitations identified in this setup.

\subsubsection{Orbital-cycle experiment}

For the second on-orbit configuration, the analysis window of the LLM agent was extended to 90 minutes to align with the ISS's orbital period. This duration was chosen based on the observation that conditions in the immediate subsequent orbit are typically similar to those in the current one, allowing the system to leverage data from the previous window (orbit) as a proxy for estimating the optimal $\alpha$ value for the next one. By synchronizing the reasoning cycle with these predictable orbital dynamics the setup aimed to mitigate the temporal mismatches identified in the short-cycle experiment, ensuring that $\alpha$ recommendations remain relevant at the time of application while still accommodating the LLM's inference latency.

In this configuration, the agentic system demonstrated substantial improvements across key metrics, achieving significantly longer episode durations and a marked reduction in thermal violations compared to the RL-only baseline. CPU utilization showed a 20.13\% increase on average, reflecting enhanced resource efficiency without compromising thermal safety(Table~\ref{table:iss_results_v2}).

\begin{table}[h]
\caption{Representative results from on-orbit orbital-cycle experiments aboard the ISS.}
\label{table:iss_results_v2}
\centering
\small
\resizebox{\columnwidth}{!}{
\begin{tabular}{|l|r|r|}
    \hline \hline
    \multicolumn{1}{|c|}{\textbf{Metric}} & 
    \multicolumn{1}{c|}{\makecell{\textbf{Baseline}\\ (Mean $\pm$ Std. Dev)}} & 
    \multicolumn{1}{c|}{\makecell{\textbf{Agentic}\\ (Mean $\pm$ Std. Dev)}} \\
    \hline
    Avg. Episode Duration & $449.77 \pm 295.69$ & $\mathbf{1555.09 \pm 823.57}$ \\
    \hline
    Thermal Violations    & $66 \pm 24.25$ & $\mathbf{22.33 \pm 9.07}$ \\
    \hline
    Avg. CPU Usage (\%)   & $40.17 \pm 17.70$ & $\mathbf{48.27 \pm 11.49}$ \\
    \hline \hline
\end{tabular}
}
\end{table}
The enhanced performance can be attributed to the improved alignment between the LLM's decision cycle and the orbital thermal dynamics, which allowed for more timely and contextually appropriate $\alpha$ adjustments. This temporal synchronization enabled the agentic modulation to promote stable exploratory or exploitative behaviors that better anticipated the cyclical environmental shifts, reducing the risk of outdated recommendations. As a result, the system not only minimized thermal violations through proactive adaptation but also extended operational episodes by maintaining temperatures closer to the threshold without exceeding it. These findings validate the potential of orbit-aligned reasoning windows in overcoming latency constraints, providing a more robust framework for agentic supervision in dynamic LEO environments.

\subsection{Comparative Discussion}
The discrepancy between ground and on-orbit results underscores a critical limitation of agentic LLM-based systems in flight conditions: when the environment's dominant timescale is comparable to or shorter than the LLM inference latency, semantic supervision can degrade performance. 

In the ground experiments, ASTREA excelled, achieving a 67.12\% increase in episode duration and a 58.48\% reduction in thermal violations over the baseline during the first 4 hours, with stable CPU utilization, due to gradual thermal changes aligning with the LLM's decision cycle. 

However, the short-cycle on-orbit experiment (15-minute window) showed a 19.12\% decrease in episode duration and a 24.24\% increase in violations, as rapid thermal shifts rendered $\alpha$ recommendations outdated, often triggering inappropriate exploration during conservative phases.

The orbital-cycle experiment (90-minute window), aligning the reasoning cycle with the ISS orbit, reversed these trends, yielding a 245.75\% increase in episode duration and a 66.17\% reduction in violations, with a 20.13\% increase in average CPU utilization. This success highlights that synchronizing the LLM's analysis with environmental periodicity can mitigate latency issues, enabling effective supervision. 

These findings affirm a key principle: LLMs are unsuitable for real-time intervention unless their latency is significantly shorter than the controlled process's dynamics. The results provide practical design guidelines, emphasizing asynchronous, orbit-aligned architectures for future LEO autonomy systems.


%% file: chapters/05_General_Discussion.tex
\section{Implications for Future Space Autonomy}
\label{sec:discussion}

The experimentation confirms that the integration of linguistic models for semantic supervision of other processes is feasible in flight-qualified equipment within the transformative New Space era. Agentic systems are thus positioned as a viable and, given their inherent characteristics, transformative approach to integrating control logic into traditional subsystems. 

The deployment of ASTREA demonstrates how agentic systems can serve as supervisory layers for traditional control architectures. By combining linguistic reasoning with adaptive control, ASTREA exemplifies a scalable approach to onboard autonomy, capable of supporting fault detection, mission planning, and subsystem coordination.

Given the analytical capabilities that extend beyond rigid rule-based systems, it is reasonable to anticipate that the integration of agentic systems aboard spacecraft will become the norm rather than the exception in the near future. While these systems may not assume complete mission control, they are well-positioned to assist other subsystems, collaborate with additional LLM agents or other agent types as proposed in this study, or simply analyze and characterize emergency levels from satellite telemetry to provide early warnings to ground control systems. Furthermore, an agentic system could autonomously plan onboard analyses of sensor data using specialized models or corroborate telemetry through alternative methods utilizing other sensors typically present on the satellite, mirroring the approach an engineer would take when confronted with anomalous telemetry readings.

This capability assumes even greater significance in remote exploration missions, providing the ability to perform contextual analysis while applying pseudo-reasoning to decision-making processes, thereby conferring a high degree of automation and autonomy to space systems.

This type of onboard analysis and automation would be impractical to achieve with other technologies. Even implementing a fraction of such analytical capability would require extensive development and validation periods under simulated extreme conditions, inevitably excluding parameters not considered during the design and implementation phases of such software. This limitation eliminates any capacity for improvisation, which is often precisely what is needed in highly dynamic and unpredictable environments.

However, incorporating the most powerful linguistic models into flight-qualified hardware remains technically unfeasible at present. This constraint necessitates maintaining short context lengths in prompts and structuring them in such a way that smaller models, compatible with current flight hardware capabilities, can properly contextualize the information received in the prompt and respond appropriately.

Another critical consideration in the deployment of such systems is the grounding problem. As Schoepp S. et al. expose, this manifests in the integration of LLM-agents with other systems that expect more structured inputs, such as reinforcement learning agents, requiring the adaptation of natural language generated by the LLM-agent into executable actions for the RL-agent \cite{schoepp2025evolving}. One approach to address or mitigate these grounding challenges, as implemented in this study's use case, is to employ the well-established Tool Use design pattern. This approach constrains the model to utilize tools that generate concrete actions, enabling the model itself to perform the necessary adaptation as part of its analytical process for response generation.

%% file: chapters/06_Conclusions.tex
\section{Conclusions and Future Work}
\label{sec:conclusions}

This work demonstrated the feasibility of deploying agentic systems on flight-qualified hardware to provide an upper layer of semantic supervision for autonomous edge control. By integrating a quantized Large Language Model (LLM) with a reinforcement learning (RL) agent, ASTREA showed that even compact linguistic models, despite their limited contextual capacity, can effectively complement adaptive controllers under strict hardware constraints. With appropriate optimization techniques, the hybrid architecture delivered measurable improvements in the thermal control agent’s performance, validating the potential of LLM-based reasoning to enhance long-term decision-making in spaceborne systems.

A single LLM agent was employed, exhibiting noticeable response-time limitations that would become even more critical in multi-agent configurations, where inference delays could generate bottlenecks, an aspect warranting further investigation. Likewise, the absence of dedicated accelerators on the current platform forced all inference to run on the CPU alongside other processes. The use of space-qualified accelerators could fundamentally transform performance, enabling larger models, shorter reasoning windows, and more frequent supervisory updates.

Although the LLM in this study focused on modulating the entropy coefficient ($\alpha$) of the RL model, future extensions could target other control parameters or incorporate adaptive reward shaping. Moreover, repeating the experiments with domain-specialized models in thermal management could strengthen contextual grounding and reduce dependence on prompt engineering. Overall, ASTREA establishes a concrete milestone toward embedding agentic reasoning in flight systems, demonstrating the technical readiness of semantic autonomy for next-generation space missions.